\documentclass{article}

\usepackage{microtype}
\usepackage{graphicx}
\usepackage{subfigure}
\usepackage{booktabs} % for professional tables

\usepackage{hyperref}

\usepackage[accepted]{icml2024}

\usepackage{amsmath}
\usepackage{amssymb}
\usepackage{mathtools}
\usepackage{amsthm}

\usepackage{pgfplots}

\usepackage[capitalize,noabbrev]{cleveref}

\theoremstyle{plain}

\theoremstyle{definition}

\theoremstyle{remark}

\usepackage[textsize=tiny]{todonotes}
\usepackage{multirow}

\icmltitlerunning{GPTree: Towards Explainable Decision-Making via LLM-powered Decision Trees}

\begin{document}

\twocolumn[
\icmltitle{GPTree: Towards Explainable Decision-Making via LLM-powered Decision Trees}



\icmlsetsymbol{equal}{*}

\begin{icmlauthorlist}
\icmlauthor{Sichao Xiong}{yyy,comp}
\icmlauthor{Yigit Ihlamur}{comp}
\icmlauthor{Fuat Alican}{comp}
\icmlauthor{Aaron Ontoyin Yin}{comp}
%\icmlauthor{Firstname3 Lastname3}{yyy}
\end{icmlauthorlist}

\icmlaffiliation{yyy}{Department of Computer Science, University of Oxford, Oxford, United Kingdom}
\icmlaffiliation{comp}{Vela Research, San Francisco, United States}
\icmlaffiliation{comp}{Vela Research, San Francisco, United States}
\icmlaffiliation{comp}{Vela Research, San Francisco, United States}

\icmlcorrespondingauthor{Sichao Xiong}{sichao.xiong@lmh.ox.ac.uk}
\icmlcorrespondingauthor{Yigit Ihlamur}{yigit@vela.partners}
\icmlcorrespondingauthor{Fuat Alican}{fuat@vela.partners}
\icmlcorrespondingauthor{Aaron Ontoyin Yin}{aaron@vela.partners}

\icmlkeywords{Machine Learning, ICML}

\vskip 0.3in
]



\printAffiliationsAndNotice{}  % leave blank if no need to mention equal contribution

\begin{abstract}
Traditional decision tree algorithms are explainable but struggle with non-linear, high-dimensional data, limiting its applicability in complex decision-making. Neural networks excel at capturing complex patterns but sacrifice explainability in the process. In this work, we present GPTree, a novel framework combining explainability of decision trees with the advanced reasoning capabilities of LLMs. GPTree eliminates the need for feature engineering and prompt chaining, requiring only a task-specific prompt and leveraging a tree-based structure to dynamically split samples. We also introduce an expert-in-the-loop feedback mechanism to further enhance performance by enabling human intervention to refine and rebuild decision paths, emphasizing the harmony between human expertise and machine intelligence. Our decision tree achieved a 7.8\% precision rate for identifying “unicorn” startups at the inception stage of a startup, surpassing gpt-4o with few-shot learning as well as the best human decision-makers (3.1\% to 5.6\%).
\end{abstract}
\section{Introduction}
Thanks to their explainability, decision trees are among the most intuitive and popular methods in machine learning (second only to linear regression). Their tree-like structure enables users to easily follow the decision-making process, with each split representing an interpretable rule based on the input data. This transparency is valuable in domains such as the Venture Capital (VC) industry where the stakes are high for each investment decision. In practice, however, decision trees often underperform when faced with non-linear, high-dimensional datasets and are inherently unsuitable for text-rich and multi-modal datasets. Thus, our work to extend decision trees is motivated by this fundamental question: \textit{how to incorporate LLMs?}

\begin{figure}[t!]
    %\centering
    \begin{tikzpicture}
        \begin{axis}[
            width=8.3cm, height=7.5cm,
            xbar,
            symbolic y coords={{Best GPTree\\model}, {GPTree\\w/expert}, GPTree, {Tier-1 seed\\funds}, gpt-4o, {Indexing\\strategy}},
            ytick=data,
            y tick label style={font=\small,  inner sep=2pt, align=right},
            nodes near coords,
            xlabel={Precision (\%)},
            ylabel={},
            xmin=0, xmax=21,
            bar width=0.5cm,
            enlarge y limits=0.15,
            ]
            \addplot coordinates {(17.9,{Best GPTree\\model}) (7.8,{GPTree\\w/expert}) (7.2,GPTree) (5.6,{Tier-1 seed\\funds}) (3.1,gpt-4o) (1.9,{Indexing\\strategy})};
        \end{axis}
    \end{tikzpicture}
    \vspace{-15px}
    \caption{Comparison of Different Models/Methods}
\end{figure}
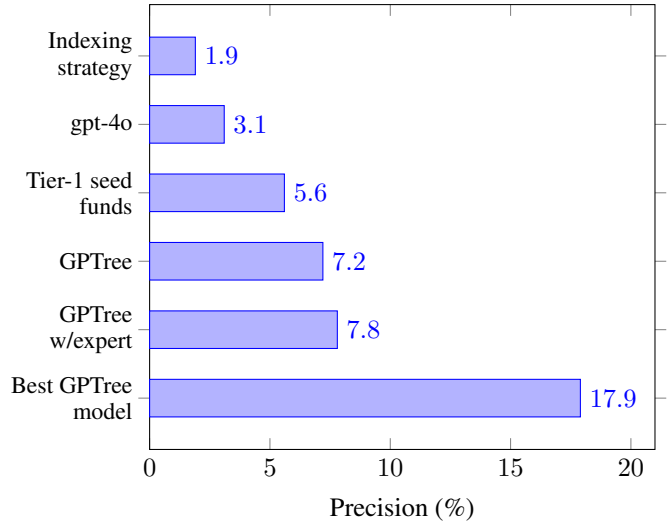

In recent years, Large Language Models (LLMs) have emerged as powerful tools capable of capturing the intricacies of natural language: models such as gpt-4o and gpt-1o preview have showcased exceptional capabilities in advanced reasoning and multi-modal tasks. However, LLMs are often seen as ``black boxes" due to their elusive architectures. In addition, prompt engineering techniques like chain-of-thought and tree-of-thought, combined with extensive prompt chaining, are frequently required to produce accurate and contextually relevant responses. Not only does this require considerable human intervention and expertise but is also prone to trial-and-error procedures in crafting prompts that provide meaningful outputs. Therefore, we address another question: \textit{how to design a robust and explainable approach that minimizes human intervention while maintaining high performance?}

In this paper, we present GPTree, a novel framework that combines the explainability of decision trees with the advanced reasoning capabilities of LLMs, also incorporating an efficient expert-in-the-loop feedback system. 
\begin{figure*}[t]
    \centering
    \includegraphics[width=\textwidth]{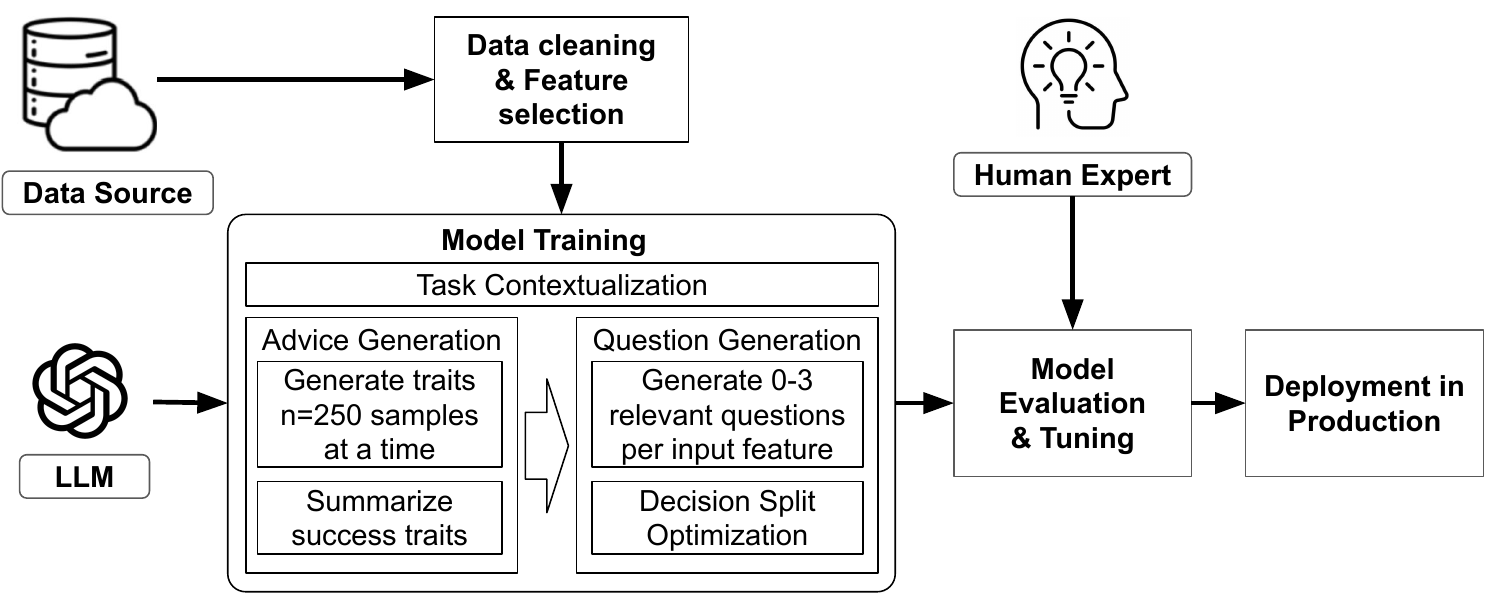}
    \caption{GPTree pipeline}
    \label{fig:pipeline}
\end{figure*}

In summary, our work makes the following contributions:
\begin{itemize}
    \item We introduce an LLM-powered decision tree model to dynamically split samples using a combination of LLM inference, code-based and clustering nodes, giving users the full explainability of traditional decision trees along with the flexibility of working with unstructured text and potentially multimodal datasets.
    \item We eliminate the need for feature engineering and prompt chaining, instead replacing it with our expert-in-the-loop feedback mechanism. This further enhances performance by enabling a human expert to refine and rebuild decision paths post-training, leveraging the cooperation potential between human expertise and machine intelligence.
    \item We conduct a comprehensive empirical evaluation of our approach within the VC landscape, where explainable decision-making is essential. Our experimental analysis demonstrates the effectiveness of our approach at identifying “unicorn” startups in comparison to human decision-makers within the industry, based on data collected from over 115K US-based companies founded more than 8 years ago.
\end{itemize}

\section{Dataset}
\subsection{Founder Success Dataset}
This dataset contains 9,892 founder profiles collected and enriched from publicly available data sources as well as paid data sources such as Crunchbase. These profiles have been drawn from companies that were founded between 2010 and 2016 and have raised between \$100K and \$4M in funding. A founder is classified as successful if his/her company achieves a valuation of more than \$500M through an IPO (initial public offering), is acquired for more than \$500M, or holds a large funding round of more than \$500M. Of the 9,892 founders, 978 (9.9\%) are categorized as successful, while 8,914 are considered unsuccessful.

\section{GPTree}

\begin{figure}[h]
    \centering
    \includegraphics[width=0.48\textwidth]{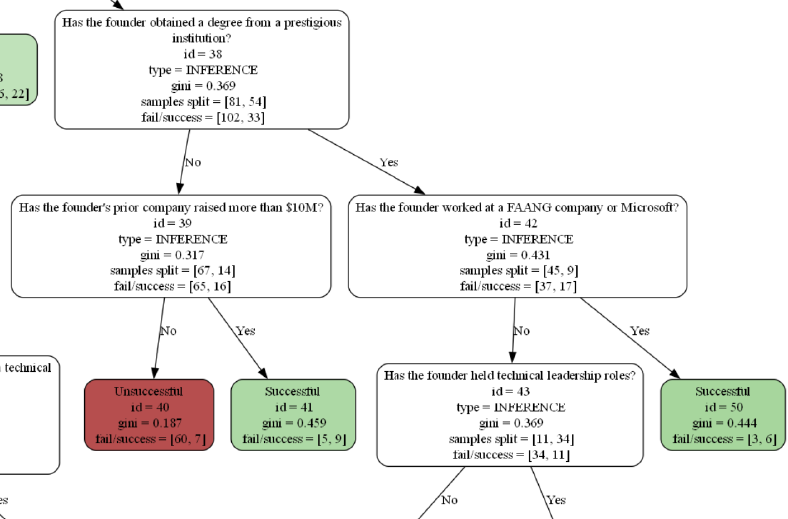}
    \caption{Decision Tree example}
    \label{fig:example}
\end{figure}
GPTree is an automated framework that combines the reasoning and generative capabilities of the latest foundation models \cite{openai2024gpt4technicalreport} with the explainability and robustness of decision trees, delivering intelligent and adaptive decision-making. 

The process consists of several key stages, as illustrated in Figure~\ref{fig:pipeline}: \textit{Task Contextualization}, \textit{Insight Generation}, \textit{Question Candidate Generation}, \textit{Decision Split Optimization}, and \textit{Expert Refinement}. While we currently use GPT-4o-mini as our backend to query, this approach can easily be adapted as more advanced models become available, ensuring scalability and improved accuracy over time. 

In Figure~\ref{fig:example} we illustrate sample output generated by our approach. 
We detail each stage in the ensuing sections.
\subsection{Task Contextualization}
GPTree starts by taking a task-specific string as user input to be used as context for the LLM. Task contextualization boosts performance as follows:
\begin{itemize}
    \item Enhanced Task Understanding: Parsing a task string allows the LLM to function as a human expert, allowing it to utilize appropriate context from its training corpus, leading to more meaningful and actionable insights.
    \item Improving Focus: The LLM can prioritize specific information relevant to the task, reducing unnecessary output and improving the precision of its response.
    \item Consistency: Task-specific prompts ensure that repeated queries on similar data or questions yield consistent outputs, which is crucial for high-stakes tasks like data analysis or decision-making.
\end{itemize}

As a concrete example, if our task is to distinguish successful founders, the following instruction might be used:

\texttt{"Imagine you are a VC analyst. Analyze the given data of successful founders and identify common features or success patterns. Provide a concise summary of these common characteristics/traits."}

\subsection{Insight Generation}
It is well known that LLMs can be inherently ``lazy" \cite{Tang_2023}, often opting for shortcuts when tasked to execute multiple steps simultaneously.  We observed that performance declines significantly when they are asked to perform analysis on successful founders and generate appropriate question candidates in a single instance. This is especially pronounced when dealing with large datasets, as they often exceed the context window of the LLM, leading to superficial analysis and outcomes, which are not representative of the entire dataset.

To address this issue, we first generate advice for the LLM based on the input data in a systematic manner. By iterating through the samples with $batch\_size=250$ (depending on the size of the input data) and producing concise summaries of the common characteristics of the successful samples, we make sure that the model focuses on relevant features without being overwhelmed by the volume of data. Subsequently, another instance of the LLM synthesizes these summaries into a cohesive list of insights. This refined list is then passed onto the question candidate generation process.

\subsection{Question Candidate Generation}
Unlike traditional decision trees, which have a finite number of input values and therefore a finite set of candidate conditions to test, natural language inputs introduce an unbounded question space. To make the problem more tractable, we incorporate advice generated from the previous stage along with the task context and ask the LLM to generate 0-3 questions per input feature as candidate questions as follows:

\texttt{"Your task is to distinguish successful from unsuccessful ones by generating precise questions based on a given Dataframe containing various founder features. You are to act as a decision node in a decision tree, formulating questions that can help distinguish successful founders..."}

At this stage we classify each question into one of the three types: INFERENCE, CODE, or CLUSTERING. For CODE-type questions, we generate a Python lambda function that takes the DataFrame as input and outputs a pair of DataFrames corresponding to the 'yes' and 'no' instances of the question.

While CODE nodes benefit from fast execution and fully deterministic outputs, their reliability is contingent on the correctness/suitability of the code generated by current models and still remains a challenge. For categorically-based clustering, we set a threshold for the maximum branching factor. And if the number of categories exceeds this value, then an LLM instance will further group categories based on similarity so that it satisfies the given threshold. Due to the limited number of categorical input features in our dataset, for experimentation purposes we restrict ourselves to only INFERENCE nodes.

\subsection{Decision Split Optimization}
Given the list of question candidates, we compute the samples in the split and check whether or not they meet the minimum number of required samples. Then, for each valid split, we compute the individual Gini impurity and use a greedy strategy to select the question which minimizes the overall weighted Gini at that particular node. 

This heuristic allows us to increase the homogeneity of the data splits and ensures that we have the best possible separation. The weighted Gini impurity formula is as follows
\begin{equation}
G_{\text{weighted}} = \sum_{i=1}^{k} \frac{n_i}{N} G_i
\end{equation}
where 
$k$ is the number of child nodes after the split (typically 2),
$n_i$ is the number of samples in the $i$-th child node,
$N$ is the total number of samples across all child nodes (i.e., $N = \sum_{i=1}^{k} n_i$),
$G_i$ is the Gini impurity of the $i$-th child node.

\begin{equation}
G_i = 1 - \sum_{j=1}^{C} p_{ij}^2
\end{equation}
where $C$ is the number of classes,
$p_{ij}$ is the proportion of samples in the $i$-th child node that belongs to class $j$.

The tree is then constructed recursively, terminating when the condition of minimum samples per node is met, max depth is reached or a particular node becomes ``pure".
\subsection{Expert Refinement}
Post training, human experts can use their knowledge to fine-tune the model on a separate validation set by rebuilding and refining decision paths using the following implemented features:
\begin{itemize}
    \item Collapse a node
    \item Rebuild subtree at a node
    \item Q\&A with the samples
    \item Remove trivial nodes
\end{itemize}
As an example the following string could be used as advice to rebuild a specific subtree.

\texttt{Consider if the founder worked at big tech companies such as Google, Microsoft, Apple and Facebook/Meta. Consider if the founder worked at a public tech company (NASDAQ).  Consider if the founder has studied at a top 20 ranked university based on QS World University Ranking 2023....}

The sensitivity hyperparameter classifies the leaf nodes as successful or unsuccessful based on the condition $\text{sensitivity} \geq \frac{\#\text{success samples}}{\#\text{total samples}}$. We optimize the model by selecting the sensitivity which maximizes the $F_{0.5}$ score. The model is then run on the test set to obtain the final set of metrics. 
\section{Experiments}
\subsection{Evaluation Metrics and Baselines}
Due to limited resources, we prioritize precision over recall, as ensuring that an invested company is successful is much more important than capturing all successful companies; therefore, we evaluate our models using $F_{0.5}$ score. In other contexts such as medical diagnosis, it may be more appropriate to consider $F_1$ or $F_2$. The following section illustrates the baselines that we used.

\subsection{Industry benchmarks}
The metrics in Table~\ref{table:seed_fund_comparison} show the precision of human decision-makers based on data collected from over 115K US-based companies founded between 2014 and 2016. All the companies considered have raised more than \$100K with Indexing Strategy representing the proportion of founders who are classified as successful. 
\begin{table}[h!]
\centering
\begin{tabular}{l c}
\toprule
Seed Fund & Outlier Rate (\%)\\
\midrule
Indexing Strategy & 1.9 \\
Y Combinator & 3.2 \\
9 Tier-1 VCs & \textbf{5.6} \\
\bottomrule
\end{tabular}
\caption{Seed funds baseline}
\label{table:seed_fund_comparison}
\end{table}

Note that the metrics mentioned in the remainder of this section are based on the 9.9\% founder success rate present in our dataset and thus should be scaled down 5.5$\times$ to match the 1.9\% outlier rate in the industry.

\subsection{Vanilla gpt-4o}
As a baseline, we use the following prompt to generate vanilla predictions without any prompt engineering techniques or decision trees. This allows us to evaluate GPT-4's inherent ability to distinguish between successful and unsuccessful founders, relying solely on general knowledge and context embedded within its training corpus.

\texttt{"You are an expert in venture capital tasked with distinguishing successful founders from unsuccessful ones. All founders under consideration are sourced from LinkedIn profiles of companies that have raised between \$100K and \$4M in funding. A successful founder is defined as one whose company has achieved either an exit or IPO valued at over \$500M or raised more than \$500M in funding."}

\begin{table}[h!]
\centering
\begin{tabular}{l c c c c}
\toprule
Model & Accuracy & Precision & Recall & $F_{0.5}$\\
\midrule
gpt-4o-mini & \textbf{87.1} & 14.4 & 6.2 & 12.9 \\
gpt-4o & 83.8 & \textbf{15.7} & \textbf{14.5} & \textbf{15.5} \\
\bottomrule
\end{tabular}
\caption{Vanilla prompting baseline}
\label{table:vanilla_comparison}
\end{table}

As shown in Table~\ref{table:vanilla_comparison}, gpt-4o exhibits a 1.3 percentage point improvement in precision and, notably, has more than twice the recall of gpt-4o-mini. A precision of $15.7\%$ represents $1.59\times$ the base proportion (9.9\%) of positive samples. This demonstrates GPT-4o's ability to make meaningful inferences even without specialized techniques, providing a foundation for further enhancements with our decision tree approach.

\subsection{Few-shot prompting}
Based on the few-shot prompting concept introduced by \cite{brown2020languagemodelsfewshotlearners}, we refine our previous vanilla prompt by providing four labelled founders: two successful and two unsuccessful. The goal is to help the model recognize similar contextual signals from the founders' profiles when making future predictions. We expect the model to generalize from a small number of examples, potentially improving performance over the vanilla baseline by aligning the task more closely with the desired outcome.

\begin{table}[h!]
\centering
\begin{tabular}{l c c c c}
\toprule
Model & Accuracy & Precision & Recall & $F_{0.5}$\\
\midrule
gpt-4o-mini & 88.6 & 15.2 & 3.37 & 8.9 \\
gpt-4o & \textbf{88.8} & \textbf{16.2} & \textbf{6.9} & \textbf{12.7} \\
\bottomrule
\end{tabular}
\caption{Few shot-prompting baseline}
\label{table:few-shot_comparison}
\end{table}

Table~\ref{table:few-shot_comparison} exhibits a significant drop in recall across both models, compensated by a marginal increase in precision. This reduction in recall is likely due to the limited variety of examples, which may cause the models to overfit to the specific instances provided, restricting their ability to generalize across unseen data.

\subsection{GPTree}
In Table~\ref{table:averaged_results} we have the averaged results from the test set, with the individual fold results of the cross-validation along with the fine-tuning results displayed in Table~\ref{tab:results}. Overall, we significantly outperform all human decision-makers. Aside from the reported cross-validation values, the proprietary model at Vela (trained without cross-validation) achieves 10$\times$ the baseline and 2.29$\times$ the cross-validated GPTree with a precision of 17.9\%, post-scaling and after normalizing that to the real world distribution. We utilized a different train-test split and dedicated a significant amount of effort to refine the tree based on advice from VCs at Vela Partners. For the purpose of academic studies, we have omitted that in this section given that there was no cross-validation performed on that model due to the intensive time required for fine-tuning. 

\begin{table}[h]
\centering
\begin{tabular}{l c c c c}
\toprule
Model & Accuracy & Precision & Recall & $F_{0.5}$\\
\midrule
GPTree & 87.6 & 37.3 & \textbf{27.1} & \textbf{33.4} \\
w/ expert & \textbf{88.7} & \textbf{40.8} & 23.4 & 33.2 \\
\bottomrule
\end{tabular}
\caption{Averaged GPTree results}
\label{table:averaged_results}
\end{table}

\begin{table*}[ht]
\centering
\begin{tabular}{l|c|c c c|cc c c c}
\toprule
 & \multicolumn{4}{c}{Validation} & \multicolumn{4}{c}{Test} \\
\cmidrule(lr){2-5} \cmidrule(lr){6-9}
Fold & Sensitivity& Precision & Recall & $F_{0.5}$ & Accuracy & Precision & Recall & $F_{0.5}$ \\
\midrule
\multirow{2}{*}{1} & 0.32 (0.31) & 48.9 (48.6) & 30.5 (33.8) & 43.6 (44.7) & 89.5 (89.8) & 50.0 (52.2) & 28.4 (33.7) & 43.4 (\textbf{47.0})\\
 & 0.37 (0.37) & 57.1 (\textbf{60.9})& 26.9 (33.7)& 46.7 (\textbf{52.4})& 89.2 (89.6)& 50.0 (53.3)& 24.9 (26.8)& 41.6 (44.5)\\
\multirow{2}{*}{2} & 0.33 (0.31)& 51.4 (51.1)& 27.4 (34.1)& 43.7 (46.5)& 90.3 (90.4)& 49.0 (50.0)& 26.3 (33.2)& 41.8 (45.4)\\
 & 0.33 (0.38)& 49.0 (53.4)& 26.3 (28.9)& 41.8 (45.7)& 89.6 (89.8)& 51.4 (53.0)& 27.4 (29.8)& 43.7 (45.9)\\
\multirow{2}{*}{3} & 0.30 (0.30)& 28.4 (32.2)& 21.9 (20.2)& 26.8 (28.8)& 88.4 (88.5)& 42.7 (41.4)& 32.5 (25.7)& 40.2 (36.9)\\
 & 0.31 (0.32)& 44.7 (50.5)& 30.6 (23.8)& 40.9 (41.2)& 87.9 (89.1)& 27.9 (31.4)& 19.7 (14.8)& 25.8 (25.6)\\
\multirow{2}{*}{4} & 0.28 (0.28)& 40.0 (41.7)& 25.3 (26.8)& 35.8 (37.5)& 88.3 (88.2)& 42.2 (42.0)& 23.0 (25.8)& 36.2 (37.3)\\
 & 0.28 (0.28)& 42.2 (42.0)& 23.0 (25.8)& 36.2 (37.3)& 88.7 (88.9)& 40.0 (41.7)& 25.3 (26.8)& 35.8 (37.5)\\
\multirow{2}{*}{5} & 0.24 (0.27)& 27.2 (32.3)& 40.8 (31.6)& 29.2 (32.2)& 84.4 (87.8)& 33.1 (40.2)& 48.5 (\textbf{35.9})& 35.4 (39.3)\\
 & 0.30 (0.29)& 43.0 (45.1)& 28.2 (29.1)& 38.9 (40.7)& 88.3 (88.3)& 33.6 (34.5)& 18.9 (19.9)& 29.1 (30.1)\\
\multirow{2}{*}{6} & 0.35 (0.34)& 36.3 (58.7)& 20.2 (14.8)& 31.3 (36.8)& 88.7 (89.8)& 35.7 (42.9)& 17.9 (9.2)& 29.8 (24.7)\\
 & 0.31 (0.29)& 36.1 (35.5)& 26.5 (33.7)& 33.7 (35.1)& 88.0 (87.1)& 30.5 (30.5)& 23.5 (31.1)& 28.8 (30.6)\\
\multirow{2}{*}{7} & 0.33 (0.34)& 43.7 (53.0)& 21.3 (19.7)& 36.1 (39.6)& 88.0 (88.4)& 35.4 (38.3)& 13.0 (10.7)& 26.4 (25.3)\\
 & 0.26 (0.25)& 30.9 (34.8)& 29.3 (26.5)& 30.6 (32.7)& 86.8 (88.2)& 30.1 (33.3)& 35.4 (31.5)& 31.1 (32.9)\\
\multirow{2}{*}{8*} & 0.31 (0.28)& 32.8 (47.1)& 24.7 (9.0)& 30.8 (25.5)& 87.5 (89.7)& 32.2 (\textbf{55.0})& 18.4 (5.3)& 28.0 (19.2)\\
 & 0.12 (0.12)& 26.1 (32.1)& 49.5 (30.6)& 28.8 (31.8)& 79.2 (85.7)& 19.9 (22.6)& 43.4 (24.2)& 22.3 (22.9)\\
\multirow{2}{*}{9} & 0.37 (0.35)& 50.0 (51.2)& 24.4 (24.2)& 41.2 (41.8)& 90.0 (90.1)& 42.1 (43.0)& 21.9 (21.9)& 35.5 (36.0)\\
 & 0.48 (0.48)& 47.0 (48.1)& 21.3 (21.3)& 37.9 (38.5)& 91.1 (\textbf{91.2})& 50.6 (51.9)& 22.5 (22.5)& 40.5 (41.2)\\
\multirow{2}{*}{10*} & 0.21 (0.37)& 24.2 (42.2)& 36.0 (10.7)& 25.9 (26.5)& 83.1 (89.6)& 25.7 (34.4)& 37.2 (5.6)& 27.4 (17.0)\\
 & 0.22 (0.22)& 26.8 (29.1)& 36.7 (\textbf{35.2})& 28.3 (30.2)& 84.4 (84.4)& 24.3 (23.7)& 34.8 (33.1)& 25.9 (25.1)\\
\midrule
Avg. & - & 39.3 (44.5)& 28.5 (25.7)& 35.4 (37.3)& 87.6 (88.7)& 37.3 (40.8)& 27.1 (23.4)& 33.4 (33.2)\\
\bottomrule
\end{tabular}
\caption{Validation and Test Results}
\label{tab:results}
\end{table*}

\section{Training details}
This section describes the GPTree training strategy. We implement several parameters/features present in scikit-learn DecisionTreeClassifier.  

\textbf{Initialization and hyperparameters.} To prevent overfitting, we train our decision trees using a maximum tree depth of 18 and a minimum of 31 (0.5\%) samples per leaf node. These hyperparameters are chosen to ensure a balance between model complexity and generalization performance. We perform summarization with a batch size of 250 samples.

\textbf{Cross-Validation.} To demonstrate the robustness of our approach, we perform 5-fold cross validation, considering all ten possible combinations of three folds (i.e., \( \binom{5}{3} \)) for training, with two different ways to assign validation and test folds. We generate 10 distinct trees, and report the averaged performance metrics over all 20 partitions. This reduces the variance and ensuring that the reported performance is not dependent on specific fold assignments. We cross-validate with a validation in order to perform expert-in-the-loop.  

\textbf{Optimization.} Similar to traditional decision trees, we use the Gini impurity to measure the quality of splits at each node. At each iteration, the optimizer selects the question candidate, which minimizes the weighted Gini impurity of the resulting child nodes. This ensures that each split creates the most homogeneous subtrees possible. Post training, we select the sensitivity on the validation set, which optimizes the $F_{0.5}$ score (favouring precision over recall). 

\textbf{Fine-tuning expert-in-the-loop.} We fine-tuned the trained decision tree with expert in the loop on the validation set. It includes rebuilding subtrees at certain nodes and collapsing the tree at certain nodes. Then we test the fine-tuned tree on a held out test set.

\textbf{Training time.} A typical training of 6000 samples using gpt-4o-mini typically takes 10 hours on a single 8-core CPU costing about \$30 in API usage. This also scales based on the number of input features that are present, which affects the number of question candidates at each stage. For comparison, gpt-4o is 40\% faster for question answering but 33x more expensive. For the full cross-validation process, which involves training 10 distinct trees, the total training time is approximately four days followed by two days of work for the expert refinement. The main bottleneck lies in the speed of API calls. However, there is potential for further acceleration through the use of more CPU cores for parallel processing.

\section{Conclusion}
In this paper, we introduced GPTree, a novel LLM-powered decision tree model, with an efficient expert-in-the-loop mechanism, emphasizing the harmony between human expertise and machine intelligence. We presented empirical evidence demonstrating the efficacy of our approach, and highlighting its ability to enhance predictive accuracy and decision-making processes in the VC industry.

\subsection{Future work} As foundation models continue to evolve, we expect that GPTree's performance will scale, making it applicable to other domains, such as explainable healthcare diagnosis. In the future, we expect that this work can be generalized to multi-modal application including images, video and audio. 

\subsection{Limitations} While our work presents promising advancements, we recognize that it has several limitations: 1) Unreliability of CODE execution nodes; 2) Non-deterministic LLM evaluations (i.e. different phrasings of similar questions can result in different results); 3) Hallucination when there is insufficient information to answer the questions.

It is important to note that the findings presented are inherently limited to the scope and characteristics of the dataset. They may reflect biases associated with the voluntary provision, or omission, of information by the founders. Additionally, due to the nature of data acquisition through scraping techniques, there may be data quality issues that could potentially lead to inaccurate conclusions. Readers should exercise caution and consider these limitations when interpreting the results.

Although we demonstrate the effectiveness of our approach using a founder success dataset, this paper is not intended as financial advice, investment guidance, or a recommendation to engage in any financial activities. Some publicly available datasets utilized in this study have been augmented with paid data sources, which may restrict the wide distribution of the underlying enriched training dataset.

\section*{Impact Statement}
The goal of the work presented in this paper is to advance the field of Machine Learning, emphasizing a harmonic relationship between humans and machines. All datasets have been processed in strict adherence to the relevant code of conduct. 

%\section*{Acknowledgements}

\bibliography{main}
\bibliographystyle{icml2024}

%%%%%%%%%%%%%%%%%%%%%%%%%%%%%%%%%%%%%%%%%%%%%%%%%%%%%%%%%%%%%%%%%%%%%%%%%%%%%%%
%%%%%%%%%%%%%%%%%%%%%%%%%%%%%%%%%%%%%%%%%%%%%%%%%%%%%%%%%%%%%%%%%%%%%%%%%%%%%%%
% APPENDIX
%%%%%%%%%%%%%%%%%%%%%%%%%%%%%%%%%%%%%%%%%%%%%%%%%%%%%%%%%%%%%%%%%%%%%%%%%%%%%%%
%%%%%%%%%%%%%%%%%%%%%%%%%%%%%%%%%%%%%%%%%%%%%%%%%%%%%%%%%%%%%%%%%%%%%%%%%%%%%%%
%\newpage
%\appendix
%\onecolumn
%\section{You \emph{can} have an appendix here.}

%You can have as much text here as you want. The main body must be at most $8$ pages long.
%For the final version, one more page can be added.
%If you want, you can use an appendix like this one.  

%The $\mathtt{\backslash onecolumn}$ command above can be kept in place if you prefer a one-column appendix, or can be removed if you prefer a two-column appendix.  Apart from this possible change, the style (font size, spacing, margins, page numbering, etc.) should be kept the same as the main body.
%%%%%%%%%%%%%%%%%%%%%%%%%%%%%%%%%%%%%%%%%%%%%%%%%%%%%%%%%%%%%%%%%%%%%%%%%%%%%%%
%%%%%%%%%%%%%%%%%%%%%%%%%%%%%%%%%%%%%%%%%%%%%%%%%%%%%%%%%%%%%%%%%%%%%%%%%%%%%%%

\end{document}